\documentclass[letterpaper]{article} 
\usepackage{aaai2026}  
\usepackage{times}  
\usepackage{helvet}  
\usepackage{courier}  
\usepackage[hyphens]{url}  
\usepackage{graphicx} 
\usepackage{natbib}  
\usepackage{caption} 
\usepackage{soul}
\usepackage{url}
\usepackage[utf8]{inputenc}
\usepackage{caption}
\usepackage{booktabs}
\usepackage{algorithm}
\usepackage{algorithmic}
\usepackage[switch]{lineno}
\usepackage{pifont}

\usepackage{amsmath,amssymb,amsfonts}
\usepackage{textcomp}
\usepackage{multirow}
\usepackage{subfigure}
\usepackage{array}
\usepackage{colortbl}
\usepackage{xcolor}
\usepackage{mathrsfs}
\usepackage{algorithm}
\usepackage{algorithmic}

\usepackage{newfloat}
\usepackage{listings}
\DeclareCaptionStyle{ruled}{labelfont=normalfont,labelsep=colon,strut=off} 
\floatstyle{ruled}
\newfloat{listing}{tb}{lst}{}
\floatname{listing}{Listing}
\title{Recursive Visual Imagination and Adaptive Linguistic Grounding for Vision Language Navigation}
\author{
    Bolei Chen$^1$,
    Jiaxu Kang$^1$,
    Yifei Wang$^1$,
    Ping Zhong$^1$\thanks{Corresponding author.},
    Qi Wu$^2$,
    Jianxin Wang$^{1*}$
}
\affiliations{
    \textsuperscript{\rm 1}School of Computer Science and Engineering, Central South University\\
    \textsuperscript{\rm 2}Australian Institute of Machine Learning, The University of Adelaide\\

    \{boleichen, jxkang, yifeiwang, ping.zhong\}@csu.edu.cn, qi.wu01@adelaide.edu.au, jxwang@mail.csu.edu.cn
}

\usepackage{bibentry}

\begin{document}

\maketitle

\begin{abstract}
\textbf{V}ision \textbf{L}anguage \textbf{N}avigation (VLN) typically requires agents to navigate to specified objects or remote regions in unknown scenes by obeying linguistic commands. Such tasks require organizing historical visual observations for linguistic grounding, which is critical for long-sequence navigational decisions. However, current agents suffer from overly detailed scene representation and ambiguous vision-language alignment, which weaken their comprehension of navigation-friendly high-level scene priors and easily lead to behaviors that violate linguistic commands. To tackle these issues, we propose a navigation policy by recursively summarizing along-the-way visual perceptions, which are adaptively aligned with commands to enhance linguistic grounding. In particular, by structurally modeling historical trajectories as compact neural grids, several \textbf{R}ecursive \textbf{V}isual \textbf{I}magination (RVI) techniques are proposed to motivate agents to focus on the regularity of visual transitions and semantic scene layouts, instead of dealing with misleading geometric details. Then, an \textbf{A}daptive \textbf{L}inguistic \textbf{G}rounding (ALG) technique is proposed to align the learned situational memories with different linguistic components purposefully. Such fine-grained semantic matching facilitates the accurate anticipation of navigation actions and progress. Our navigation policy outperforms the state-of-the-art methods on the challenging VLN-CE and ObjectNav tasks, showing the superiority of our RVI and ALG techniques for VLN.
\end{abstract}


\begin{figure}[!t]
 \centering
 \includegraphics[width=1.0\linewidth]{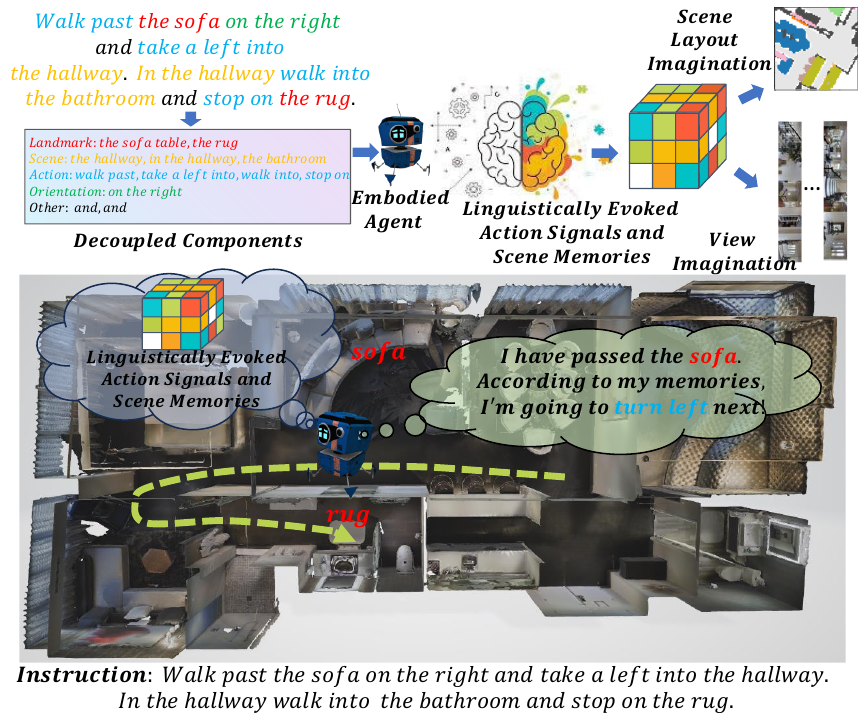}
 \caption{The VLN agent decouples an instruction into different components, including landmarks, scenes, actions, orientations, and others, which are adaptively aligned with high-level scene priors in the ISR. The pre-trained ISR can provide the necessary mindsets for VLN, including view imagination and scene layout imagination. 
 }
 \label{fig1}
\end{figure}

\section{Introduction}

Interacting with agents through natural language is a long-term goal of embodied artificial intelligence as it is potentially the most intuitive way for human-robot communication. The emerging research on \textbf{V}ision \textbf{L}anguage \textbf{N}avigation (VLN) \cite{Th2022Navigating,an2024etpnav} is along this path, which requires agents to navigate to specified object instances or remote areas in unfamiliar 3D scenes by following linguistic instructions. Existing VLN work has made great advances in \textbf{S}cene \textbf{R}epresentation (SR) \cite{wang2023gridmm,hong2023learning,an2024etpnav}, vision-language alignment \cite{cui2023grounded,cheng2022learning}, and auxiliary tasks \cite{wu2024vision,qiao2023hop+} for pre-training. They typically organize historical visual observations as structural SRs, which are further cross-modally aligned with linguistic commands to track navigation progress and enhance navigation decision-making. 

Some methods \cite{3DAware, yokoyama2024vlfm,wang2023gridmm} represent scenes by projecting raw or encoded visual features into bird's-eye-view maps or 3D feature fields to preserve fine-grained scene geometries and visual contexts. Despite promising progress has been made, these SRs provide overly detailed structural and semantic priors, posing challenges for learning accurate vision-action mappings using neural networks. Human-like agents typically establish high-level awareness of landmark semantics and spatial relationships of surrounding objects, rather than focusing on misleading geometric details that are irrelevant to navigation. For example, the agent in Fig. \ref{fig1} should focus on the sofa landmarks and the visual signals that trigger the left-turn action, rather than the objects' visual textures and the hallway's geometric structure. Research in behavioral psychology \cite{tolman1948cognitive,o1996geometric} has shown that many animals maintain spatial representations of their scenes during navigation, even if scene details are not fully stored. Inspired by this, some other methods \cite{an2024etpnav,yin2024sg} propose to abstract the environmental layouts into visual feature-based \textbf{T}opological \textbf{S}cene \textbf{R}epresentations (TSR) to facilitate linguistic grounding or balance exploration and exploitation during navigation. Although TSR refines the scene layout, TSR's nodes still store raw or encoded visual textures that are overly detailed. Moreover, TSR discards continuous semantic relations between nodes \cite{chen2023think}.

Redundant SRs can impede linguistic grounding, potentially resulting in behaviors contradicting navigation instructions. In other words, redundant scene details that are irrelevant to VLN can disrupt effective linguistic grounding, leading to ambiguous or even erroneous vision-language alignment. Current methods \cite{wang2023gridmm,hong2023learning,an2024etpnav} attempt to align instruction tokens with SRs through standard cross-modal attention techniques. In this case, it is extremely challenging to train a transformer to achieve disentanglement and match each instruction token to the correct visual feature in a redundant SR. Such an ambiguous semantic alignment impairs the agent's insight into the navigation progress and makes it easy to deviate from the correct trajectory. 

To tackle these issues, we propose a VLN policy by organizing along-the-way observations as an \textbf{I}mplicit \textbf{S}cene \textbf{R}epresentation (ISR) through \textbf{R}ecursive \textbf{V}isual \textbf{I}magination (RVI), including view imagination and scene layout imagination. Technically, we advocate modeling historical navigation trajectories (including the agent's visual sensing, poses, and navigational actions) as compact neural grids, rather than preserving explicit scene geometric details. We treat SR learning as a sequence modeling problem and train a joint state-action transformer over entire trajectories under the behavior cloning framework \cite{hu2024transforming}. Unlike classical VLN methods \cite{HAMT,wang2023dual}, the number of neural grids in our ISR is a hyperparameter that does not grow with trajectory length or scene scale. Therefore, the number of ISR tokens input to our model is fixed, which does not increase the computational cost. Then, the learned ISR is densely aligned with navigation commands via a novel \textbf{A}daptive \textbf{L}inguistic \textbf{G}rounding (ALG) technique to make the vision-language matching clear.

To derive navigation-friendly high-level scene priors from an ISR, RVI motivates agents to focus on the regularity of visual transitions and semantic scene layouts while ignoring irrelevant visual contexts. In particular, view imagination motivates agents to learn the distribution of future visual frames while enhancing their sensitivity to historical visual changes. Due to the inherent uncertainty in future frame prediction and the diversity of navigational actions, a single current frame can generate multiple potential futures. Therefore, our VLN agent is encouraged to summarize the regularity of visual signal changes instead of deterministically rendering future visual features. Scene layout imagination is designed to enhance the agent's insights into the surrounding landmark semantics and their relative positional relations. Therefore, our core idea is to explicitly endow the agent with the thinking necessary for VLN: \textbf{(1) recalling the past and predicting the future and (2) imagining the current semantic layout of the surroundings.}

Research in brain science \cite{sokolov2017cerebellum,vargha1997differential} has shown that the cerebellum and hippocampus regulate motion and memory recall through neural structures and feature representations, respectively. Inspired by this, the ALG technique is proposed to adaptively align ISR's neural grids with different linguistic components for vision-language matching.  For example, \textit{left turn} action signals and \textit{sofa} associated situational memories should be governed by separate neural grids, as shown in Fig. \ref{fig1}. To realize this idea, the agent first decouples a navigation instruction into different components, including landmarks, scenes, actions, and orientations, through syntactic analysis. Then, a self-supervised learning method is proposed to adaptively align these components with appropriate action signals or scene memories at the positional and semantic levels. 

During experiments, sufficient comparative studies reflect that our approach incorporating RVI and ALG achieves state-of-the-art performance on two VLN tasks. Adequate ablation studies validate the effectiveness of the individual modules of our method. In general, the main contributions of this paper are as follows: \textbf{(1)} Two novel RVI techniques are designed for ISR learning that can empower agents with the essential thinking for VLN. \textbf{(2)} A novel ALG technique is proposed to motivate the agent to adaptively activate different action signals or scene memories based on different linguistic components. \textbf{(3)} Sufficient comparative and ablative studies on challenging VLN tasks demonstrate the superiority of our method. The experimental code will be publicly available after anonymous review.

\section{Related Work} \label{section2}

\textbf{Scene Representation for VLN.} Effective SRs are essential for the long-sequence decision-making and vision-instruction alignment of VLN. Early efforts \cite{dang2022unbiased,tan2024self} typically employ recurrent neural networks to model SR as a fixed-size feature vector, which may be inefficient in modeling sophisticated visual features and capturing the long-term feature dependence in historical trajectories. Due to the strong expression power of transformer \cite{hu2024transforming}, transformer-based models \cite{qiao2023hop+,wu2024vision,cui2023grounded,wang2023gridmm,lin2022multimodal} have manifested their potential in VLN. Among them, architecture enhancement methods \cite{lin2022multimodal,chen2021history,hong2021vln} consider how to apply the powerful transformer structure to VLN under the reinforcement learning framework, facilitating more precise modeling of scenes. Trajectory optimization methods \cite{wang2023gridmm,qiao2023hop+,cui2023grounded,wu2024vision} treat VLN tasks as sequence modeling problems and train joint state-action models over entire trajectories under the behavior cloning framework. 

Alternatively, some other methods \cite{wang2023dual,an2023bevbert,wang2023gridmm} achieve SR by projecting encoded visual features into egocentric semantic maps or topological graphs, which exhaustively retain the visual contexts and scene geometries. Although these methods achieve promising results, their SRs contain redundant information. We argue that SR should adequately represent the high-level scene-understanding mindsets required for VLN, rather than providing agents with excessive and misleading scene details. Inspired by the trajectory optimization methods \cite{wu2024vision,ehsani2024spoc}, we propose an ISR by modeling historical observations as compact neural grids. Unlike existing methods \cite{wang2023gridmm,wu2024vision,chen2021history,hong2021vln}, we condense and refine the valuable historical information before feeding it into the cross-modal fusion module. In other words, the ISR is learned to emphasize the agent’s insights into high-level visual signals and semantic scene layouts, which is distinct from existing SR modeling. 

\textbf{Linguistic Grounding for VLN.} Fine-grained linguistic grounding is critical for instruction-following action prediction and VLN progress tracking. However, existing methods \cite{wang2023gridmm,an2024etpnav,georgakis2022cross,an2023bevbert} coarsely align all instruction tokens with the SR at the sentence level, which impairs the agent’s insight into the navigation progress. Some other studies \cite{wu2024vision,qiao2023hop+} adopt auxiliary tasks to sequentially align historical observations with instructions during the pre-training phase. However, the positional and semantic alignments between historical observations and instruction tokens are still ambiguous. To mitigate these issues, alternative methods \cite{cui2023grounded,cheng2022learning} decouple navigation instructions into actions and landmarks and match them with entities in the panoramic images at a fine-grained level. However, given the diversity of scenes and the complexity of instructions, it is inadequate to bridge the vision-language gap using only navigational actions and entity landmarks. 

To address the above issues, we propose to decouple a navigation instruction into different components, including landmarks, scenes, actions, and orientations. Then, an ALG technique is proposed to achieve dense alignment between the linguistic components and the ISR at the positional and semantic levels, respectively. The ALG technique allows VLN agents to evoke different episodic memories adaptively according to different linguistic components.

\section{Preliminaries} \label{section3}

\textbf{Problem Definition.} In this work, we address the VLN tasks in 3D indoor scenes, where the agents are required to reach specified remote regions or object instances. In particular, we focus on two practical settings: VLN in \textbf{C}ontinuous \textbf{E}nvironments (VLN-CE) \cite{krantz2020beyond} and \textbf{Object}-goal \textbf{Nav}igation (ObjectNav) \cite{Th2022Navigating} tasks in continuous scenes, where the agents should take low-level navigational actions. The action space consists of a set of parameterized discrete actions, e.g., \textit{Forward (0.25m)}, \textit{Turn Left/Right (15$^\circ$)}, and \textit{Stop}. Both VLN-CE and ObjectNav utilize the Habitat simulator \cite{ramakrishnan2021habitat} to render RGB and depth observations based on the MatterPort3D (MP3D) \cite{chang2017matterport3d} dataset. In addition, the agents can receive noiseless 3-DoF pose data $(x, y, \theta)$, including 2D position and 1D orientation. At timestep $t$, the VLN agent can observe panoramic RGB images $\mathcal{R}_t = \{ I_{t,k}^{rgb}\}_{k=1}^K$ and depth images $\mathcal{D}_t = \{ I_{t,k}^{depth}\}_{k=1}^K$ of its current location, which both contain $K$ single view images. The VLN agent also receives an instruction with $L$ words for each episode, which are embedded as $X = \{x_i\}_{i=1}^L$. The ObjectNav agent can observe one single RGB image $I_t^{rgb}$ and one single depth image $I_t^{depth}$. In each episode, the ObjectNav agent is given a target category $c_{target}$ specified by a semantic label (e.g., a toilet). To facilitate the learning of a unified VLN framework, ObjectNav's goal is converted to “\textit{Please navigate to [$c_{target}$] and stay within 1 m of it.}” by using a fixed instruction template. Unless otherwise stated, we default to introducing our method under the VLN setup.

\begin{figure}[!t]
 \centering
 \includegraphics[width=1.0\linewidth]{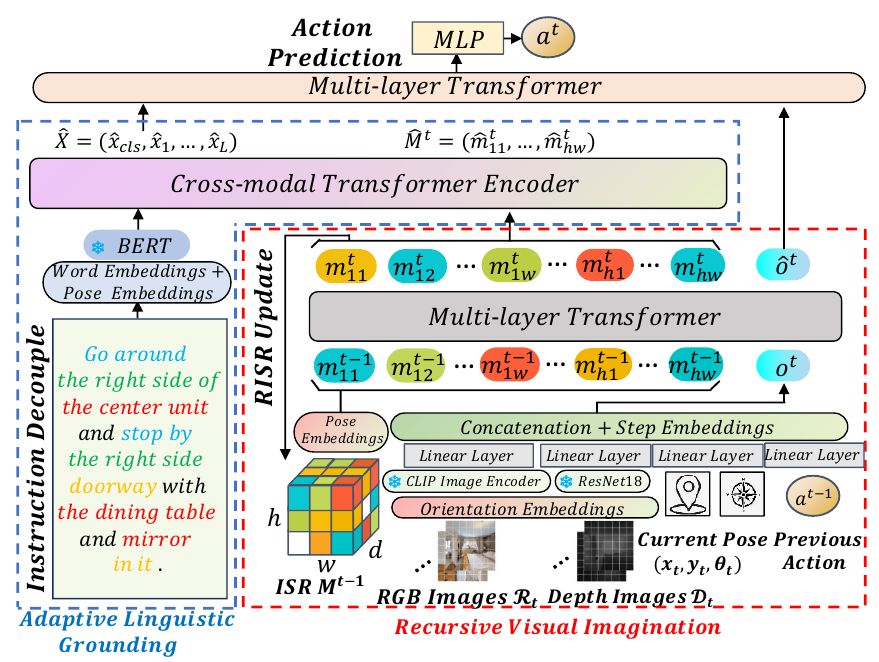}
 \caption{An illustration of our VLN policy with RVI (Fig. \ref{fig3}) and ALG (Fig. \ref{fig4}). Our method treats SR learning as a sequence modeling problem and trains a joint state-action transformer over entire trajectories.}
 \label{fig2}
  \vspace{-0.3cm}
\end{figure}

\textbf{ISR Initialization and Updating.} At timestep $t$, the agent's observations specifically include the panoramic RGB-D images $\{\mathcal{R}_t,\mathcal{D}_t\}$, the pose $(x_t,y_t,\theta_t)$, and the previous navigation action $a^{t-1}$, as shown in Fig. \ref{fig2}. Following existing work \cite{wang2023dreamwalker, an2024etpnav, wang2023gridmm}, we first perform orientation embedding for each view of the panoramic image. Then, the pre-trained CLIP ResNet50 \cite{radford2021learning} and the ResNet18 pre-trained in PointNav \cite{wijmans2019dd} are used to encode the individual RGB view $I_{t,k}^{rgb}$ and depth view $I_{t,k}^{depth}$, respectively. Notably, the visual encoders stay frozen to make the training efficient. The agent's current pose is converted into a vector $(x_t,y_t,sin\theta_t,cos\theta_t)$ before encoding. Four different linear layers are used to project the visual embeddings, the pose vector, and the previous action into the same dimension. All the features are concatenated and further added a sinusoidal positional embedding of timestep $t$ to obtain the current observation feature $o^t$.

Our ISR summarizes the historical images until timestep $t$ as neural grids $M^t=[m^t_{ij}]_{h \times w}$ with $h \times w$ grids. Each grid is a $d$-dimensional feature vector $m^t_{ij} \in \mathbb{R}^d$ whose position with respect to the center is designated $[i-h/2,j-w/2]$. As each episode starts, the neural grids $M^0$ are initialized using their positions $m^0_{ij}=w^0_m+MLP([i-h/2,j-w/2])$,
where $w^0_m \in \mathbb{R}^d$ is a learnable embedding. At each timestep, the neural grids are updated given the new observation $o^t$ with a differentiable function. Given the effectiveness of transformers in sequential modeling and VLN \cite{HAMT}, a multi-layer transformer is employed to achieve interactions among neural grid-based situational memories. We first perform positional embedding for neural grids to enhance the geometry alignment between the neural grids and the observation. Then, all the neural grids and $o^t$ are concatenated as tokens which are fed to the transformer, as shown in Fig. \ref{fig2}. 

Notably, unlike the voxels for 3D scene reconstruction, we introduce the concept of a ``grid” to emphasize the relative positional encoding of ISR. In the following section, we expect agents to predict local semantic maps during RVI, which requires inferring the relative positional relations between high-level semantics. In addition, we expect the grids with different positions to be aligned with the corresponding instruction components during ALG. This is inspired by the fact that the hippocampus and cerebellum, which have different relative positions in the brain, are responsible for memory and movement, respectively.



\section{Methodology} \label{section4}
\subsection{Recursive Visual Imagination} \label{IVA}

\begin{figure}[!t]
 \centering
 \includegraphics[width=1.0\linewidth]{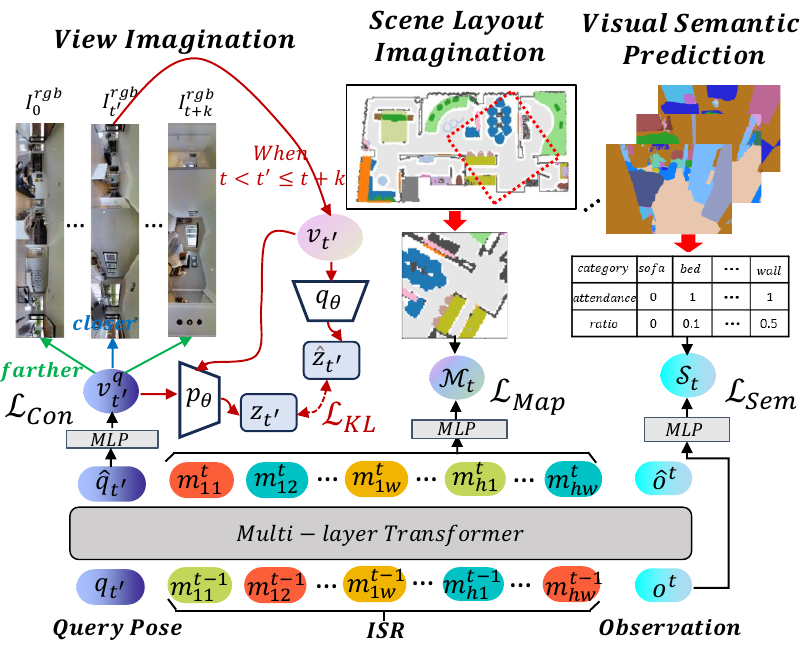}
 \caption{An illustration of RVI, including view imagination, scene layout imagination, and visual semantic prediction.}
 \label{fig3}
  \vspace{-0.3cm}
\end{figure}

To derive high-level scene priors from ISR, RVI motivates agents to focus on the regularity of visual transitions and semantic scene layouts while ignoring irrelevant visual contexts. As shown in Fig. \ref{fig3}, RVI specifically includes \textbf{V}iew \textbf{I}magination (VI), \textbf{S}cene \textbf{L}ayout \textbf{I}magination (SLI), and \textbf{V}isual \textbf{S}emantic \textbf{P}rediction (VSP).

Given a query pose, VI motivates the agent to evoke the corresponding situational memory from ISR or learn the regularity of future visual transitions. At timestep $t$, we randomly sample a query pose $\{x_{t'},y_{t'},\theta_{t'}\}$ and the corresponding RGB panoramic image $\mathcal{R}_{t'}$ from a VLN trajectory, where $t' \in [0,t+k]$. Then, a frozen pre-trained CLIP ResNet50 and a linear layer are utilized to encode $\mathcal{R}_{t'}$ and the query pose as $v_{t'}$ and $q_{t'}$, respectively. As shown in Fig. \ref{fig3}, $q_{t'}$ is fed into the multi-layer transformer along with $M^{t-1}$ and $o^t$ to query visual features about pose $\{x_{t'},y_{t'},\theta_{t'}\}$ from ISR. Notably, we only aim to extract potential features related to the query pose from the ISR, without expecting $q_{t'}$ to affect the ISR updating. Therefore, an attention masking operation is employed to prevent $M^{t-1}$ and $o^t$ from paying attention to $q_{t'}$. The output pose embedding is fed into an \textbf{M}ulti-\textbf{L}ayer \textbf{P}erception (MLP) to predict the visual feature $v_{t'}^q$. To enhance the agent's sensitivity to historical visual changes, we use a contrastive loss to clarify the correspondence between the poses and visual features by pushing $v^q_{t'}$ and $v_{t'}$ closer to each other and moving $v^q_{t'}$ away from visual features at other locations in the trajectory:
 \begin{flalign} \footnotesize
\begin{aligned}
\label{eq3}
&\ \mathcal{L}_{Con} = \dfrac{1}{T}\sum_{t=0}^{T}-log \dfrac{exp(sim(v_{t'}^q,v_{t'})/\tau)}{\sum_{i=1}^{t+k}exp(sim(v_{t'}^q,v_i)/\tau)}, &
\end{aligned}
\end{flalign}
where $\tau$ is a softmax temperature scaling parameter and $sim(\cdot, \cdot)$ corresponds to the cosine similarity. 

Notably, by setting the value of $k>0$, the agent is motivated to imagine visual features for the future $k$ timesteps at specific locations. To make the agent further summarize the regularity of future visual transitions, we aim to learn the distribution of future frames conditional on the current frame, rather than deterministically rendering future visual features. In particular, we employ two MLPs $p_{\vartheta}$ and $q_{\vartheta}$ to approximate the learned prior distribution $z_{t'} \sim p_{\vartheta}(z_{t'}|v_{t'}^q)$ and the posterior distribution $\hat{z}_{t'} \sim q_{\vartheta}(\hat{z}_{t'}|v_{t'}^q, v_{t'})$ that captures future uncertainty, respectively.  We make the prior distribution to be closer to the posterior distribution by minimizing the KL divergence, which not only enables the agent to fantasize about the future but also makes the future variable more predictable. In summary, the loss function for visual imagination is as follows:
\begin{equation} \footnotesize
\begin{aligned}
\label{eq4}
\mathcal{L}_{VF} = \mathcal{L}_{Con} + \beta KL[q_{\vartheta}(z_{t'}|v_{t'}^q, v_{t'})||p_{\vartheta}(z_{t'}|v_{t'}^q)],
\end{aligned}
\end{equation}
where $\beta$ is a loss scale hyperparameter. When $0 < t'\le t$, $\beta = 0$, otherwise $\beta = 0.5\ (t < t'\le t+k)$.

SLI is designed to enhance the agent's insights into the surrounding landmark semantics and the relative positional relationships among them. Technically, an MLP is used to predict egocentric local semantic maps $\{\mathcal{M}^t\}_{t=0}^T$ from ISR, where $\{\mathcal{M}^t\}_{t=0}^T$ is pre-generated from the MP3D dataset, as shown in Fig. \ref{fig3}. Please see the supplementary material for more details of $\mathcal{M}^t \in \mathbb{R}^{H \times W}$. A \textbf{B}inary \textbf{C}ross-\textbf{E}ntropy (BCE) loss is used to measure the SLI error:
\begin{flalign} \footnotesize
\begin{aligned}
\label{eq5}
\mathcal{L}_{Map} = \frac{1}{T}\sum_{t=0}^T BCE(Linear(M^t),\mathcal{M}^t).
\end{aligned}
\end{flalign}
To boost VI and SLI's focus on scene semantics, VSP is used as an auxiliary task to enhance the sensitivity of the observation encoding component to visual semantics. Technically, VSP is achieved to predict the existence of each object category and the ratio occupied by the objects in the current view (if they are present) based on the observation $o^t$, as shown in Fig. \ref{fig3}. We obtain the ground-truth labels from the MP3D training scenes and use the BCE loss $\mathcal{L}_{Sem}$ to measure the VSP errors. Please see the supplementary material for the data collection details for pre-training.

\subsection{Adaptive Linguistic Grounding} \label{IVB}

\begin{figure}[!t]
 \centering
 \includegraphics[width=1.0\linewidth]{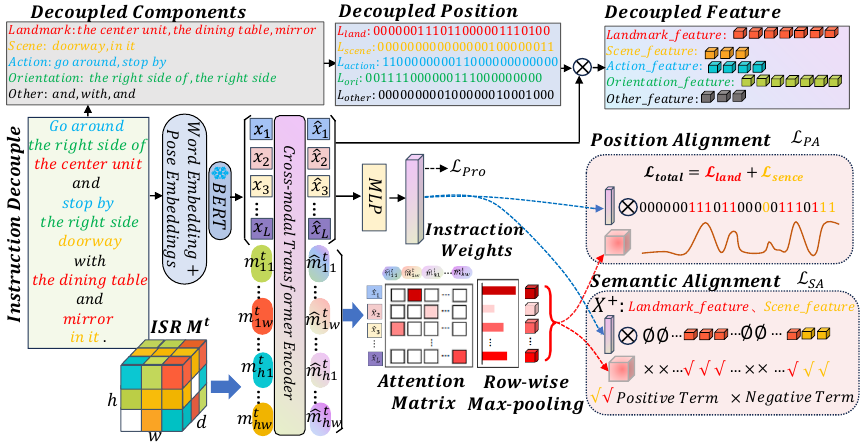}
 \caption{An illustration of ALG, including instruction decoupling, VLN progress tracking, and linguistic alignment.}
 \label{fig4}
  \vspace{-0.3cm}
\end{figure}

\textbf{\textit{Instruction Decoupling.}} Human beings can wisely focus on instruction-related landmarks in the scene and scene-related orientations in the instructions when performing VLN tasks. To emulate such abilities, we propose to decouple the instruction into different components, which are independently and adaptively aligned with ISR's neural grids, producing more discriminative and clear vision-language matching. Technically, we follow the existing work \cite{wu2019unified} to parse the instructions grammatically and decouple the instructions into five semantic components: landmarks, scenes, actions, orientations, and others. Particularly, we generate the position labels $L_{land}$, $L_{scene}$, $L_{action}$, $L_{ori}$, and $L_{other}$ for the component’s associated words by setting each component’s word positions to 1 and the rest to 0, as shown in Fig. \ref{fig4}. Given that large language models \cite{achiam2023gpt} can potentially solve this issue better, we report the related experimental results in the supplementary material. In addition, by dot-multiplying the cross-modal fused word tokens $\{ \hat{x}_i\}_{i=0}^L$ with the position labels, we derive the textual features of the decoupled components $\{\tilde{x}_i\}_{0 < i \leq L}$. Notably, the decoupled textual features, as a result of cross-attention, implicitly contain information about the global instruction and ISR while preserving the original textual features. That is, feature decoupling produces individual features while keeping the global context.

\textbf{\textit{VLN Progress Tracking.}} Since VLN's decision-making is progressive, the agent needs to track the navigation progress and explicitly align the already executed instruction components, rather than the entire instruction, with the ISR. As shown in Fig. \ref{fig4}, an MLP is used to map the cross-modal fused tokens $\hat{X}=\{ \hat{x}_1,...,\hat{x}_L\}$ to instruction weights $W_t=[\omega_1^t,...,\omega_L^t]$, which assign higher attention to the already executed instruction components. The training target $d_t$ of progress tracking is defined as the normalized distance from the current viewpoint to the goal, i.e., the target will be 1 at the beginning and closer to 0 as the agent approaches the goal. We employ a mean squared loss $\mathcal{L}_{Pro}$ to supervise the training of the progress tracking module.

\textbf{\textit{Position and Semantic Alignments.}} Before performing the ALG, we need to specify which neural grids are aligned with which components in the instruction. To this end, we propose to treat the attention matrix of the last cross-modal attention layer as an affinity matrix to match the neural grids and instruction components (as shown in Fig. \ref{fig4}), since it is learned to adaptively measure the semantic similarity between the tokens \cite{AME}. Such an idea has two benefits: \textbf{(1)} No additional matching algorithms are required. \textbf{(2)} Such a design facilitates the agent to learn neural grid's adaptive attention to different instruction components when the model parameters are updated. Specifically, we first perform row-wise max-pooling on the attention matrix to obtain each language token's most attentive neural grid $\{\tilde{m}^t_i\}_{0 < i \leq L}$. Note that $i \leq L$ since multiple language tokens pay attention to the same neural grid. Fig. \ref{fig4} shows an example of ISR actively and adaptively focusing on landmarks, scenes, i.e., positionally and semantically aligning $\{\tilde{m}^t_i\}_{0 < i \leq L}$ with the landmark and scene components in the instruction. Those tokens that do not actively pay attention to landmarks and scenes are forced to match other instruction components, i.e., actions, orientations, and others. For brevity, only the ALG technique for landmark and scene alignment shown in Fig. \ref{fig4} is detailed below.

Position alignment aims to closely match the distribution of linguistically modulated ISR with the text distribution of navigation instructions. The ground-truth text distribution of landmarks and scenes is obtained by element-wise summing the position labels of the associated decoupled text components, i.e., $L_{total}=L_{land}+L_{scene}$. In practice, we dot-multiply $L_{total}$ and $W_t$ to produce a ground-truth text distribution with navigational progress awareness, as shown in Fig. \ref{fig4}. The process of position label prediction is as follows:
 \begin{flalign} \footnotesize
\begin{aligned}
\label{eq6}
\hat{L}_{total} = Softmax(MLP(Mean([\tilde{m}_0^t,...,\tilde{m}_i^t]))),
\end{aligned}
\end{flalign}
where $Mean(\cdot)$ denotes averaging over the neural grids. We employ a BCE loss $\mathcal{L}_{PA}$ to supervise the training of the position alignment. Semantic alignment aims to match semantically similar neural grids with instruction components and keep away the dissimilar ones from both through contrastive learning. The semantic alignment loss is defined as follows:
 \begin{flalign} \scriptsize
\begin{aligned}
\label{eq7}
&\ \mathcal{L}_{SA} = \dfrac{1}{|X^+|}\sum_{\tilde{x}_i \in X^+}-log \dfrac{exp(\alpha_+ \ast (\overline{m}^{\top}\tilde{x}_i/\tau))}{\sum_{j=1}^{l}exp(\alpha_{-} \ast (\overline{m}^{\top}\tilde{x}_j/\tau))}, &
\end{aligned}
\end{flalign}
where $X^+ = \{\tilde{x}_i\}_{0<i \leq L}$ denotes the text features corresponding to the landmark and scene components, as shown in Fig. \ref{fig4}. $l$ denotes the number of tokens in $X^+$ and $\overline{m} = Mean([\tilde{m}_0^t,...,\tilde{m}_i^t])$. $\tau$ is a temperature scaling parameter. $\alpha_{+}$ and $\alpha_{-}$ are the weights of positive term (landmarks and scenes) and negative term (actions, orientations, and others), respectively. Conversely, we can also utilize the ALG technique in practice to make agents actively and adaptively focus on action and orientation components in the instruction. Those tokens that do not actively pay attention to actions and orientations are forced to align with other instruction components, i.e., landmarks, scenes, and others. Please see the supplementary material for the performance of this variant.

\subsection{Pre-training and Fine-tuning for VLN} \label{IVC}
In the pre-training phase, we train the agent using a large number of pre-collected trajectories in the behavioral cloning framework \cite{hu2024transforming}. A cross-entropy loss with inflection weighting \cite{wijmans2019embodied} is employed for action prediction, which gives higher weights for actions different from the previous one:
 \begin{flalign} \footnotesize
\begin{aligned}
\label{eq8}
\mathcal{L}_{Action} = \frac{1}{T}\sum_{t=0}^T -(1+\gamma\mathbf{1}_{a^{\ast}_t \neq a^{\ast}_{t-1}}log(p(a^{\ast}_t))).
\end{aligned}
\end{flalign}
The total loss $\mathcal{L}_{total}$ in the pre-training phase is denoted as:
 \begin{flalign} \footnotesize
\begin{aligned}
\label{eq9}
\mathcal{L}_{total} = \mathcal{L}_{Action}+\beta(\mathcal{L}_{VF}+\mathcal{L}_{Map}+\mathcal{L}_{Sem})+\\
\lambda(\mathcal{L}_{Pro}+\mathcal{L}_{PA}+\mathcal{L}_{SA}),
\end{aligned}
\end{flalign}
where $\beta$ and $\lambda$ are weighting parameters. Furthermore, the Dagger technique \cite{ross2011reduction} is used to fine-tune the pre-trained models to address the distribution discrepancy between the offline training data and the target policy. Fine-tuning fundamentally differs from the pre-training phase that employs expert demonstration paths, as it involves novel data acquisition via exploration. Please see the supplementary materials for more details.

\renewcommand\arraystretch{0.8} 
\begin{table}[t] \scriptsize
\begin{center}
\setlength{\tabcolsep}{0.4mm}{
\begin{tabular}{l | c c c | c c c }
\bottomrule[1.3pt]
\multirow{2}{*}{\textbf{Method}} &
 \multicolumn{3}{c}{\textbf{Val Unseen}}  & \multicolumn{3}{c}{\textbf{Test Unseen}}\\
\cline{2-7}
 & \textbf{OSR}$\uparrow$ & \textbf{SR}$\uparrow$ & \textbf{SPL}$\uparrow$ & \textbf{OSR}$\uparrow$ & \textbf{SR}$\uparrow$ & \textbf{SPL}$\uparrow$  \\
\hline
CM$^2$ \cite{georgakis2022cross} & 42 & 34 & 28 & 39 & 31 & 24 \\
WS-MGMap \cite{chen2022weakly} & 48 & 39 & 34 & 45 & 35 & 28 \\
GELA \cite{cui2023grounded} & 59 & 48 & 41 & 57 & 46 & 40\\
GridMM \cite{wang2023gridmm} & 61 & 49 & 41 & 56 & 46 & 39 \\
Ego$^2$-Map \cite{hong2023learning}  & - & 52 & 46 & 56 & 47 & 41 \\
DREAMWALKER \cite{wang2023dreamwalker} & 59 & 49 & 44 & 57 & 49 & 44 \\
ETPNav \cite{an2024etpnav} & 65 & 57 & 49 & 63 & 55 & 48 \\
Zhang et.al. \cite{zhang2024narrowing} & - & 58 & 49 & - & 56 & 48 \\
\hline
Ours & \textbf{67} & \textbf{59} & \textbf{50} & \textbf{64} & \textbf{57} & \textbf{50} \\
\bottomrule[1.3pt]
\end{tabular}}
\end{center}
\caption{Results on the R2R-CE dataset.}
\label{table1}
 \vspace{-0.3cm}
\end{table}

\section{Experiments} \label{section5}

\subsection{Experimental Settings and Implementation Details} 

\textbf{\textit{Datasets.}} As stated in the problem definition, we evaluate our proposed VLN strategy on the R2R-CE and Habitat ObjectNav datasets: 

\textbf{\textit{(1) R2R-CE \cite{krantz2020beyond}}} dataset comprises a total of 5,611 shortest path trajectories over 90 visually realistic scenes. To highlight our method's generalization to novel scenes, we report performance on the unseen validation (Val-Unseen) and test splits. Both splits contain episodes with novel paths and instructions from scenes that are unseen in training. An episode is successful if the stop decision is taken within 3 m of the goal position. 

\textbf{\textit{(2) ObjectNav}} experiments are performed on the MP3D dataset with the Habitat simulator. We use the standard split of 61 train / 11 val scenes with the Habitat ObjectNav dataset \cite{Th2022Navigating}, which consists of 21 goal categories. All the goals are converted to instructions such as “\textit{Please navigate to [$c_{target}$] and stay within 1 m of it.}” by using a fixed instruction template. An episode is successful if the stop decision is taken within 1 m of the object goal.

We consider these two tasks instead of the others \cite{qi2020reverie, ku2020room, anderson2018vision} because they allow agents to take low-level actions for continuous movements and are thus more practical. R2R-CE and ObjectNav require more fine-grained decisions and rely more on efficient scene representation and instruction grounding.

\textbf{\textit{Evaluation Metrics.}} There are several standard metrics \cite{an2024etpnav} for VLN evaluation, including Success Rate (SR), Oracle SR (OSR), and SR penalized by Path Length (SPL). SR (\%) gauges how often the predicted stop location is within a predefined distance from the true location. OSR (\%) determines the frequency with which any point on the predicted path is within a certain distance of the goal. SPL (\%) measures navigation effectiveness by combining the success rate with the length of the route.

\textbf{\textit{Implementation Details.}} The number of layers and attention heads of the transformers in our VLN strategy are 4 and 8, respectively. If not additionally specified, the dimensions of ISR are sized $h=w=10$ and $d=512$. $\tau$ and $k$ in VI are respectively set to 0.07 and 20. All egocentric semantic maps used in SLI have a scale of $H=W=32$ with each pixel corresponding to $20\ cm \times 20\ cm$. The $L$ in ALG is empirically set to 160 according to the length of the instructions in the R2R-CE dataset. The weights $\alpha_{+}$, $\alpha_{-}$, and $\tau$ in the semantic alignment of ALG are set to 1.0, 2.0, and 0.07, respectively. $\beta$ and $\lambda$ in Eq. \ref{eq9} are set to 0.3 and 0.5. Following existing methods \cite{wang2023dreamwalker, an2024etpnav}, we employ a waypoint predictor for the VLN task to predict long-term navigation goals. For the ObjectNav task, we directly predict low-level navigation actions end-to-end. 

For pre-training, we collect navigation trajectories based on the episodes in the training split, including visual observations, egocentric semantic maps, and semantically segmented views, please see the supplementary material for more details. The whole model is trained for 100 epochs on one NVIDIA GeForce RTX 3090 GPU using a learning rate of $1 \times 10^{-4}$ and batch size of 4. The optimizer is AdamW. For fine-tuning, our VLN policy is trained for more than 50 epochs on 4 NVIDIA GeForce RTX 3090 GPUs using a learning rate of $5 \times 10^{-5}$ and 6 threads.

\renewcommand\arraystretch{0.8} 
\begin{table}[!t] \footnotesize
\begin{center}
\setlength{\tabcolsep}{0.7mm}{
\begin{tabular}{l | c c c }
\bottomrule[1.3pt]
\multirow{2}{*}{\textbf{Method}} &
\multicolumn{3}{c}{\textbf{ObjectNav-MP3D (val)}} \\
\cline{2-4}
& \textbf{SR(\%)}$\uparrow$ & \textbf{SPL(\%)}$\uparrow$ & \textbf{DTS(m)}$\downarrow$ \\
\hline
OVRL \cite{OVRL} & 28.6 & 7.4 & -  \\
$Ego^2$-MAP \cite{EgoMap} & 29.0 & 10.6 & 5.17  \\
3D-Aware \cite{3DAware} & 34.0 & 14.6 & 4.78  \\
VLFM \cite{yokoyama2024vlfm} & 36.2 & 15.9 & - \\
ECL \cite{chen2024embodied} & 34.8 & 14.7 & 4.95  \\
SGM \cite{SGM} & 37.7 & 14.7 & 4.93  \\
T-Diff \cite{Tdiff} & 39.6 & 15.2 & 5.16  \\
SG-Nav \cite{SGNav} & 40.2 & 16.0 & -  \\
HOZ$_{e}$++ \cite{zhang2025hoz++} & 37.0 & 15.2 & \textbf{4.11}  \\
NaviFormer \cite{xie2025naviformer} & 40.1 & 15.1 & 5.19  \\
\hline
Ours & \textbf{40.9} & \textbf{17.1} & 4.68  \\
\bottomrule[1.3pt]
\end{tabular}}
\end{center}
\caption{Results on the MP3D-ObjectNav dataset (val).}
\label{table2}
 \vspace{-0.3cm}
\end{table}

\subsection{Comparison with State-of-the-art Methods}
We first conduct comparative studies between our VLN policy and the state-of-the-art methods on the R2R-CE dataset. For adequate comparisons, the baselines are diverse in terms of SR. For example, CM$^2$, GridMM, and ETPNav employ the explicit semantic grid map, visual feature field, and TSR as SRs, respectively. Ego$^2$-Map uses a self-supervised SR learning scheme based on 2D-3D contrastive learning. However, these methods share the same drawback of using only cross-attention to ambiguously align SR with instruction features at the sentence level. GELA mitigates this problem and is similar to our ALG, but it only uses contrastive learning to align visual features with the object entities in the instructions. As shown in Tab. \ref{table1}, our method achieves the best performance on both splits, reflecting the superiority of our ISR and ALG techniques. Notably, DREAMWALKER attempts to learn a world model for predicting future views to augment VLN, which is different from our visual imagination. However, DREAMWALKER requires constructing an additional TSR, which is difficult to scale to large-scale scenes. Our method overcomes this issue by using ISR to organize historical images and imagine spatio-temporal high-level semantics, thus significantly outperforms DREAMWALKER.

\renewcommand\arraystretch{0.7} 
\begin{table}[t!] \small
\begin{center}
\setlength{\tabcolsep}{1.0mm}{
\begin{tabular}{c c c | c c c | c c c }
\bottomrule[1.3pt]
\multicolumn{6}{c}{\textbf{Ablations}}&
\multicolumn{3}{c}{\textbf{Val Unseen}} \\
\hline
$\mathcal{L}_{Map}$ & $\mathcal{L}_{Con}$ & $\mathcal{L}_{KL}$ & $\mathcal{L}_{Pro}$ & $\mathcal{L}_{PA}$ & $\mathcal{L}_{SA}$  & $\textbf{OSR} \uparrow$ & $\textbf{SR} \uparrow$ & $\textbf{SPL} \uparrow$  \\
\hline
 \ding{55} & \ding{55} & \ding{55} & \ding{55} & \ding{55} & \ding{55} & 58 & 49 & 43  \\
 \checkmark & \ding{55} & \ding{55} & \ding{55} & \ding{55} & \ding{55} & 60 & 51 & 45  \\
 \checkmark & \checkmark & \ding{55} & \ding{55} & \ding{55} & \ding{55} & 62 & 52 & 45  \\
 \checkmark & \checkmark & \checkmark
 & \ding{55} & \ding{55} & \ding{55} & 63 & 53 & 47  \\
 \checkmark & \checkmark & \checkmark
 & \checkmark & \checkmark & \ding{55} & 64 & 55 & 48  \\
  \checkmark & \checkmark & \checkmark
 & \ding{55} & \checkmark & \checkmark & 63 & 54 & 46  \\
\checkmark & \checkmark & \checkmark
 & \checkmark & \checkmark & \checkmark & \textbf{67} & \textbf{58} & \textbf{50}  \\

\bottomrule[1.3pt]
\end{tabular}}
\end{center}
\caption{Ablation studies on the R2R-CE dataset.}
\label{table3}
\vspace{-0.3cm}
\end{table}

As expected, our method also achieves the best performance on the ObjectNav dataset, as shown in Tab. \ref{table2}. Similarly, our method outperforms those methods that utilize semantic grid maps (HOZ$_{e}$++ and NaviFormer), visual feature fields (VLFM), and visual representations based on self-supervised contrastive learning (OVRL, Ego$^2$-Map, and ECL). It is worth noting that T-Diff uses a trajectory diffusion technique to predict future trajectories, which is different from our idea of visual imagination. SG-Nav extracts common-sense knowledge from large language models to enhance ObjectNav, but relies on a TSR that are difficult to scale with scene size. Unlike the VLN methods in Tab. \ref{table1}, which predict navigational subgoals across multiple time steps, ObjectNav requires the agent to make navigational decisions at each time step, and thus relies more heavily on fine-grained vision-language alignment. To this end, our method has excellent visual imagination and ALG abilities, which significantly improve the ObjectNav performance.

\subsection{Ablation Studies}
We conduct ablation studies on the individual components of our method to clarify their contributions. All ablations utilize $\mathcal{L}_{Action}$ and $\mathcal{L}_{Sem}$ to ensure basic action prediction and effective observation encoding. As shown in Tab. \ref{table3}, all the RVI techniques ($\mathcal{L}_{Map}$, $\mathcal{L}_{Con}$, and $\mathcal{L}_{KL}$) can improve the VLN performance. In addition, the involvements of positional alignment $\mathcal{L}_{PA}$ and semantic alignment $\mathcal{L}_{SA}$ promote ALG, which further leads to substantial OSR, SR, and SPL boosts. Notably, $\mathcal{L}_{PA}$ and $\mathcal{L}_{SA}$ should be used in conjunction with $\mathcal{L}_{Pro}$ as the navigation process is progressive. The absence of progress tracking $\mathcal{L}_{Pro}$ will result in a significant decrease in performance.

\subsection{Diagnostic Studies and Discussion}

\begin{figure}[!t]
 \centering
\includegraphics[width=1.0\linewidth]{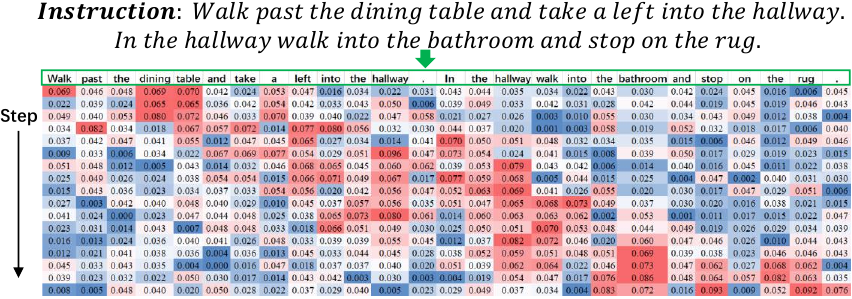}
 \caption{A visualization of how the instruction weights change with navigation progress. Different rows indicate weights at different time steps. A redder color indicates that the agent is more attentive to the corresponding words.}
 \label{fig5}
  \vspace{-0.3cm}
\end{figure}

\textbf{\textit{(1) Does the VLN progress tracking work ?}} Fig. \ref{fig5} illustrates how the instruction weights change in the process tracking module as the VLN progresses. We find that the instruction weights in the progress tracking module can reflect which part of the instruction has been executed. In addition, the instruction weights also reflect the agent's attention to the scene and landmark components of the instruction.

\begin{figure}[!t]
 \centering
\includegraphics[width=0.8\linewidth]{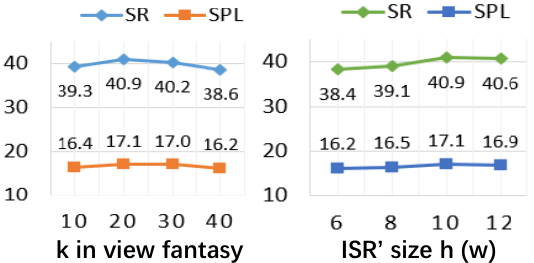}
 \caption{Illustrations of parametric studies.}
 \label{fig7}
 \vspace{-0.3cm}
\end{figure}

\textit{\textbf{(2) How much does the hyperparameters affect our method?}} Fig. \ref{fig7} illustrates the sensitivity analysis results for two key hyperparameters, i.e., the range of visual imagination ($k$), and the dimensions of ISR ($h$ and $w$). For $k$, we evaluated four cases with $k = \{10,20,30,40\}$. For $h$ and $w$, we evaluated the four cases $h=w=\{6,8,10,12\}$. We find that our method performs best when $k = 20$ and $w=h=10$. In addition, our method is insensitive to these hyperparameters and thus is robust.


\section{Conclusion} \label{section5}

This paper focuses on scene representation and instruction grounding problems in VLN tasks. For scene representation, we enable the agent's abilities to model the regularity of visual transitions and semantic scene layouts by learning an ISR, rather than retaining redundant geometric details. In other words, we advocate empowering VLN agents with two necessary abilities: \textbf{(1) recalling the past and predicting the future and (2) imagining the current semantic layout of the surroundings.} For linguistic grounding, we suggest adaptively aligning the ISR with different instruction components at the positional and semantic levels, rather than ambiguous vision-language matching. Sufficient comparative and ablation studies demonstrated our method's feasibility and superiority over existing methods. In the future, we will try to make efforts on zero-shot VLN based on multimodal large models to improve the generalization of VLN agents.


\bibliography{aaai2026}

\appendix

\begin{figure*}[!th]
 \centering
\includegraphics[width=1.0\linewidth]{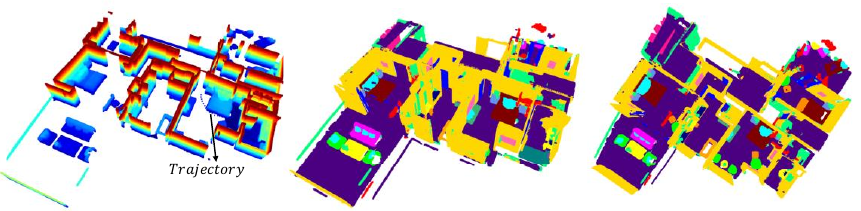}
 \caption{An example of scene and trajectory used for data collection for pre-training.}
 \label{fig8}
\end{figure*}

\section{Training Data Collection}
\begin{figure}[!t]
 \centering
\includegraphics[width=1.0\linewidth]{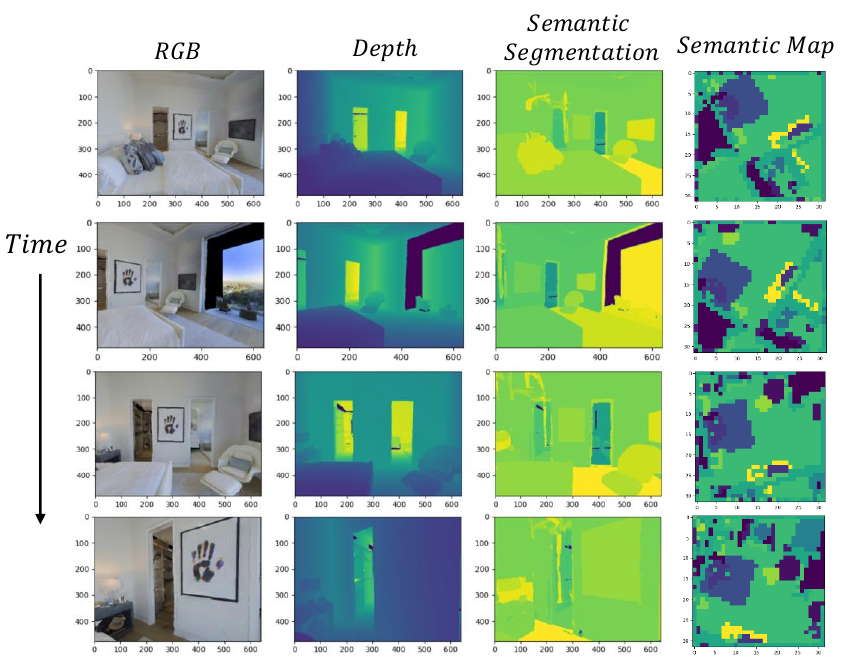}
 \caption{Examples of observation sequences collected along the trajectory in a navigation episode as shown in Fig. \ref{fig8}. Only one view per timestep is shown here.}
 \label{fig9}
\end{figure}

The data collection process used for pre-training is shown in Fig. \ref{fig8} and Fig. \ref{fig9}. The trajectories used for data collection come from the training splits of R2R-CE \cite{krantz2020beyond} and MP3D-ObjectNav \cite{chaplot2020object} datasets. In each episode, based on the MP3D scene data \cite{chang2017matterport3d}, the Habitat simulator \cite{ramakrishnan2021habitat} renders RGB, depth, and semantically segmented images at each timestep. These visual perceptions are collected as a historical observation sequence together with the agent's actions and poses. For the data generation of SLI technique, we use the semantically segmented images, depth images, and camera parameters provided by the simulator to generate a ground-truth egocentric semantic map sequence $\{\mathcal{M}^t\}_{t=0}^T$ for each VLN episode, where $\mathcal{M}^t \in \mathbb{R}^{H \times W}$. As shown in Fig. \ref{fig9}, each pixel in $\mathcal{M}^t$ stores the index of the semantic category of the corresponding position in the scene, and the MP3D dataset contains a total of 41 semantics. An index of 0 means free, otherwise it means occupied by an obstacle.

The VSP task is designed to predict the existence of each object category and the ratio occupied by the objects in the views (if they are present) based on the current observation $o^t$. We can obtain the corresponding ground-truth labels from the semantically segmented images.

\section{Model Training Details}
In the pre-training phase, we use behavioral cloning \cite{hu2024transforming} to train VLN agents. The cross-entropy loss with inflection weighting \cite{wijmans2019embodied} is employed for action prediction, which gives higher weights for actions different from the previous one:
 \begin{flalign}
\begin{aligned}
\label{eq1}
\mathcal{L}_{action\_pred} = \frac{1}{T}\sum_{t=0}^T -(1+\gamma\mathbf{1}_{a^{\ast}_t \neq a^{\ast}_{t-1}}log(p(a^{\ast}_t))).
\end{aligned}
\end{flalign}
The total loss $\mathcal{L}_{total}$ in the pre-training phase is denoted as:
 \begin{flalign} \footnotesize
\begin{aligned}
\label{eq9}
\mathcal{L}_{total} = \mathcal{L}_{Action}+\beta(\mathcal{L}_{VF}+\mathcal{L}_{Map}+\mathcal{L}_{Sem})+\\
\lambda(\mathcal{L}_{Pro}+\mathcal{L}_{PA}+\mathcal{L}_{SA}),
\end{aligned}
\end{flalign}
where $\beta$ and $\lambda$ are weighting parameters. In practice, we propose to employ $\mathcal{L}_{Action}+\beta(\mathcal{L}_{VF}+\mathcal{L}_{Map}+\mathcal{L}_{Sem})$ for the first stage of training to learn a high-quality scene representation with high-level scene priors. Then, the complete loss $\mathcal{L}_{total}$ is used for the second stage of training, which adaptively aligns the learned scene representation with the instruction components at the positional and semantic levels.

The pre-training setting can make full use of the ability of transformers to extract the optimal policy from a large amount of offline data, but it also needs to address the distribution discrepancy between the offline training data and the target policy. Therefore, the Dagger technique \cite{ross2011reduction} is used to fine-tune the pre-trained models to enhance the generalization of VLN agents, following existing works \cite{chen2022think,hong2022bridging,wang2023gridmm}. Fine-tuning fundamentally differs from the pre-training phase that employs expert demonstration paths, as it involves novel data acquisition via exploration. In particular, the model is trained with heuristic pseudo label $a_t^{pse}$, which is sampled from the distribution predicted by the agent:
 \begin{flalign}
\begin{aligned}
\label{eq3}
\mathcal{L}_{FT} = \frac{1}{T}\sum_{t=0}^T CrossEntropy(\tilde{a}_t, a_t^{pse}).
\end{aligned}
\end{flalign}

For example, a predictor \cite{hong2022bridging} is employed to generate several candidate waypoints in the VLN-CE setting. Then, the candidate waypoint nearest to the destination is used as the pseudo label $a_t^{pse}$ to encourage the agent to learn a backtracking strategy. In the initial fine-tuning phase, the waypoint closest to the destination dominates the supervision. Meanwhile, the model's uncertain decision-making drives the agent to explore the environment and reduce the exposure bias. As the model grows stronger, it will increasingly trust its own decisions so that the latter stage of the fine-tuning will be mainly supervised by the model itself.

\begin{table}[t!] \small
\begin{center}
\setlength{\tabcolsep}{2.0mm}{
\begin{tabular}{c | c c c }
\bottomrule[1.3pt]
\multirow{2}{*}{\textbf{DIA Variants}}&
\multicolumn{3}{c}{\textbf{Val Unseen}} \\
\cline{2-4} 
 & $\textbf{OSR} \uparrow$ & $\textbf{SR} \uparrow$ & $\textbf{SPL} \uparrow$  \\
\hline
 Action Priority & 65 & 55 & 47  \\
 Scene Priority  & 67 & 58 & 50  \\
\bottomrule[1.3pt]
\end{tabular}}
\end{center}
\caption{VLN performance using different ALG variants.}
\label{table4}
\end{table}

\section{Performance Evaluation of ALG Variants}

As shown in Fig. 4 in the paper, our proposed adaptive position and semantic alignments force ISR to actively focus on the landmark and scene components in the instructions, which we call $scene\ priority$. Alternatively, we can also design an action-aware ALG variant to motivate ISR to actively pay attention to the action and orientation components, which we call $action\ priority$. The comparative results in Tab. \ref{table4} quantitatively evaluate the performance of two ALG variants. We find that the focus on scene and landmark components produces more efficient VLN agents under the R2R-CE setting. In other words, agents in the R2R-CE setup are more sensitive to landmark entities and scene references.

\begin{figure}[!t]
 \centering
\includegraphics[width=1.0\linewidth]{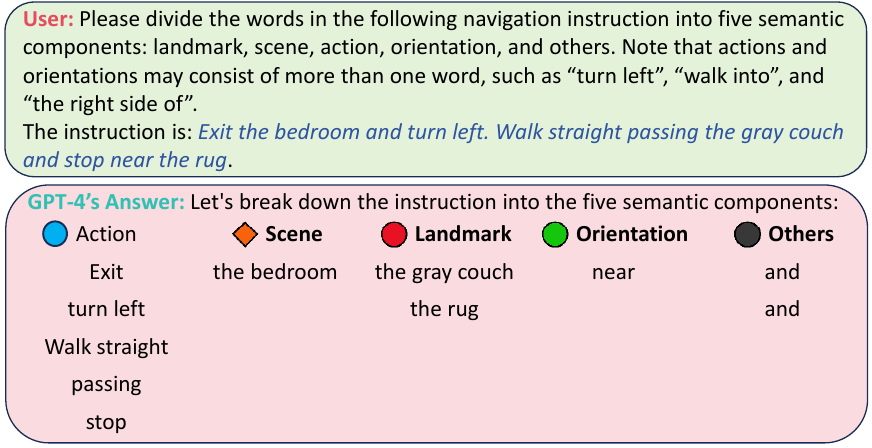}
 \caption{An illustration of semantic component division based on GPT-4.}
 \label{fig10}
\end{figure}

\begin{table}[t] 
\begin{center}
\setlength{\tabcolsep}{0.2mm}{
\begin{tabular}{l | c c c | c c c }
\bottomrule[1.3pt]
\multirow{2}{*}{\textbf{Method}} &
 \multicolumn{3}{c}{\textbf{Val Unseen}}  & \multicolumn{3}{c}{\textbf{Test Unseen}}\\
\cline{2-7}
 & $\textbf{OSR} \uparrow$ & $\textbf{SR} \uparrow$ & $\textbf{SPL} \uparrow$ & $\textbf{OSR} \uparrow$ & $\textbf{SR} \uparrow$ & $\textbf{SPL} \uparrow$  \\
\hline
w/o manual check & 66 & 58 & 49 & 63 & 57 & 50  \\
w/ manual check & 67 & 59 & 50 & 64 & 57 & 50 \\
w/ GPT-4 & 67 & 60 & 51 & 65 & 58 & 50 \\
\bottomrule[1.3pt]
\end{tabular}}
\end{center}
\caption{Effects of different instruction decoupling methods on the VLN performance on the R2R-CE dataset.}
\label{table5}
\end{table}

\section{Instruction Decoupling based on a Large Language Model}

Although performance gains have been achieved by using off-the-shelf tools \cite{schuster2015generating, wu2019unified} to decouple navigation instructions, it will inevitably lead to incorrect component divisions due to semantic ambiguities. In practice, we adjust a portion of incorrect component divisions by manually checking them. However, when more and more navigation instructions are employed to enhance the ALG, it is impractical to correct the semantic ambiguity manually. Fortunately, with the rise of large language models \cite{achiam2023gpt}, they have demonstrated language analysis and comprehension capabilities comparable to those of humans. Therefore, we prompted GPT-4 to divide navigation instructions into semantic components, including landmarks, scenes, actions, orientations, and others. An example of instruction parsing using GPT-4 is shown in Fig. \ref{fig10}, where the semantic component division is almost perfect.

In addition, we use different instruction parsing schemes to decouple the navigation instructions in the R2R-CE dataset and investigate their effects on the VLN performance, the results are shown in Tab. \ref{table5}. The first row in Tab. \ref{table5} indicates that only off-the-shelf tools are used for instruction parsing without manual checks. The second line indicates the addition of a manual check. The third line indicates directly using the components decoupled by GPT-4. The experimental results show that GPT-4-based instruction decoupling leads to better VLN performance due to the powerful language analysis capability of large language models. When manual checking is missing, the decrease in VLN performance reflects the necessity of accurate instruction decoupling for positional and semantic alignments in the ALG.

\section{More Visualization}

\begin{figure}[!t]
 \centering
\includegraphics[width=1.0\linewidth]{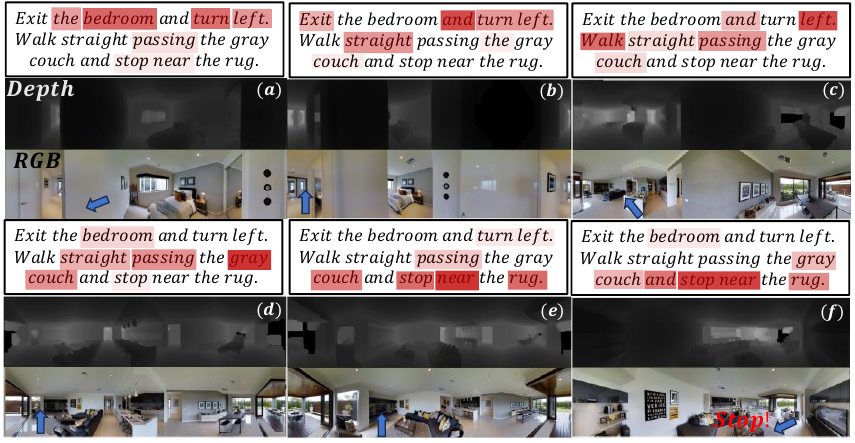}
 \caption{(a)-(f) illustrate the navigation views and process tracking during VLN. We visualize the top-6 instruction weights during process tracking in red, with darker colors having higher weights. The blue arrows indicate the navigation directions for each step.}
 \label{fig6}
\end{figure}

Fig. \ref{fig6} illustrates an example of R2R-CE with the navigation instruction “\textit{Exit the bedroom and turn left. Walk straight passing the gray couch and stop near the rug}”. The darker the base color of the words, the higher the corresponding weights and attention in Fig. \ref{fig6}. Eventually, the agent navigate to the vicinity of the rug by following the instruction.

\end{document}